\documentclass{article}

\usepackage{arxiv}

\usepackage[utf8]{inputenc} 
\usepackage[T1]{fontenc}    
\usepackage{hyperref}       
\usepackage{url}            
\usepackage{booktabs}       
\usepackage{amsfonts}       
\usepackage{nicefrac}       
\usepackage{microtype}      
\usepackage{lipsum}
\usepackage{graphicx}
\graphicspath{ {./images/} }

\title{Deepfake Detection using Biological Features: A survey}

\author{
  Kundan Patil \\
  VJTI  \\
  Mumbai\\
  \And
  Shrushti Kale \\
  VJTI  \\
  Mumbai\\
  \And
  Jaivanti Dhokey \\
  VJTI  \\
  Mumbai\\
  \And
  Abhishek Gulhane \\
  VJTI  \\
  Mumbai\\
}
\usepackage{float}
\begin{document}
\maketitle
\begin{abstract}
A deep learning-based technique called deepfake has made it easier to change or modify images and videos. In investigations and court, visual evidence is commonly employed. These pieces of evidence may now be a suspect due to technological advancements, particularly Deepfake. Photographs and movies that have been edited are incredibly lifelike and difficult to tell apart from the original. Deepfakes have been utilized to blackmail individuals, plan terrorist attacks, disseminate incorrect facts, defame individuals as well as foment political turmoil. Our study describes the history of the Deepfake to acquire a thorough understanding of the technology. Additionally, the study focuses on the development and detection of Deepfakes and their challenges based on physiological measurements. Here, biological features such as eyebrow recognition, eye blinking detection, eye movement detection, ear and mouth detection, and heartbeat detection are clearly described and a scope in this domain is proposed. A comparison of each biological feature concerning classifiers or techniques along with its key findings is discussed. Generated using the generative adversarial network (GANs) model, Deepfakes are created by iterating an actual data-based generation and verification task through two opposite deep learning models. It was easier to identify deepfakes by humans during the nascent stage of this technology owing to pixel collapse phenomena that generated visible artifacts in skin tone and the general face structure. However, with the technology’s advancement, DeepFakes have evolved to be highly indistinguishable from natural images hence it is important to review these detection methods.
\end{abstract}

\section{Introduction}
Multimedia material, such as images and videos, has grown in popularity on different social media platforms during the last several decades. One may now create a realistic-looking face that doesn't happen in reality or do a high-level realism face swap in a movie because of recent breakthroughs in deep learning-based image and video generation approaches, such as generative adversarial networks (GAN). The community refers to the latter as the Deepfake\cite{zhou2017two}. Prior until today, doing such a face swap needed expertise in areas like theatrical visual effects (VFX) and/or fast-tracking using indicators.~\cite{wodajo2021deepfake}
\vspace{3mm}
\\Therefore, the fight against Deepfakes is now in dire need of resources. There have been various attempts to recognize Deepfake movies, although Deepfake detection is a relatively newer research area ~\cite{huang2022fakelocator}. Although certain methods extensively depend on categorizing images and videos as true or fraudulent using raw image Deepfake data, others employ more traditional digital forensics methods ~\cite{chen2022pulseedit}. Moreover, if deep image synthesis technologies progress soon, such detection algorithms that just use raw pixel-domain input may develop to be less operative as Deepfake images as well as videos convert to be further convincing ~\cite{khalil2021icaps}. As a result, a completely new Deepfake detection approach is required~\cite{zhang2022cascaded}.
\vspace{3mm}
\\Deepfakes, or fake films and pictures, have caused a slew of societal problems in recent years ~\cite{ciftci2020hearts}. Deepfakes are made by iterating a real data-based creation and verification job via two opposing deep learning models utilizing the generative adversarial network (GANs) methodology~\cite{du2020towards}. Owing to the pixel collapsing phenomenon which produces false visual irregularities in the skin tone or the shape of the face or often occurring visual aberrations, Deepfake videos may be recognised with the naked eye in their early stages. While, in recent years, as technology has developed, Deepfakes have become practically impossible to tell apart from actual photographs ~\cite{yasrab2021fighting}.

\vspace{3mm}
There has been an upsurge in the incidence of indecorous use of technology as it develops. Deepfake, which produces a large number of pornographic photos of politicians and celebrities to spread propaganda, has caused a wide variety of social problems~\cite{fernandes2019predicting}. As a consequence, a lot of research has been conducted in creating a system for confirming the reliability of Deepfakes. A method for detecting the collapsing of pixels and other visible abnormalities in Deepfake has become one of the consistent assessment approaches that have become the subject of most research ~\cite{cozzolino2021id}.
\vspace{3mm}
\\However, because the creator as well as the classifier in the GANs paradigm gradually evolved to dodge such validation, using this technique has become more difficult ~\cite{ciftci2020fakecatcher}. Therefore, the usage of this approach has grown challenging because the generator as well as the discriminator in the GANs models have developed to evade such confirmation~\cite{vinay2022afmb}.

\section*{GAN (Generative Adversarial Networks)}
GAN is one of the most difficult deep learning techniques to train and apply on computational hands\cite{dong2022restricted}. There, we see two different neural networks: the generator and the discriminator. Although the generator net seems to be quite similar to the autoencoder net, we are able to produce much better outcomes because of the discriminator net's ability to exclude certain poor samples\cite{masood2022deepfakes}. Therefore, the GANs approach of Deepfake production assumes that the generator should be able to deceive the discriminator, which is another computer\cite{falahkheirkhah2022deepfake}. This is what makes these fakes more comparable to actual movies and also makes it harder for one’s eyes to recognise them\cite{almars2021deepfakes}. This method is being used by a few of the open-source projects, like Face swap-GAN.
\begin{figure}[H]
    \centering
    \includegraphics{{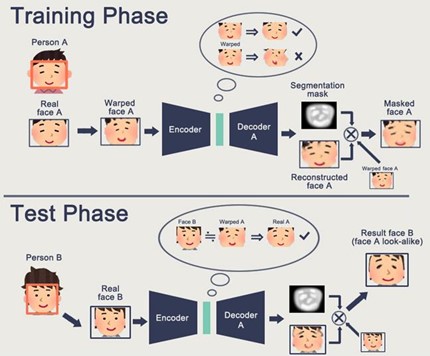}}
    \caption{Example of training GAN\cite{maksutov2020methods}}
    \label{fig:figure1}
\end{figure}
\section*{Background : An overview of Deepfake}
A politician, actress, and perhaps other celebrity's face may be switched by any other artist or people's face using the tampering or manipulation technique known as "Deepfake." This may be done with any photo or video\cite{akhtar2019comparative}. It is possible to create fake films, audio recordings or photographs that appear and sound genuine. Using massive datasets that include audio, videos or photographs as building blocks, the system constructs a model of a person speaking or doing something\cite{zendran2021swapping}. The manipulation is accomplished by using a dataset consisting of hundreds or thousands of photographs of the individual who is the subject of the investigation. Deepfake is a video and image manipulation technology that uses artificial intelligence techniques. Machine learning, an implementation of artificial intelligence, is used as it is a technique which enables a computer to adapt from the inputs of easily available information \cite{zhao2021learning}.
\vspace{3mm}
\\An application known as FakeApp was made available on Reddit, and its primary function was to walk users through the fundamental phases of the Deepfake algorithm. It is possible to make a Deepfake picture or video even with a little understanding of machine learning and programming. Deepfakes are often created for one of three reasons: to exact vengeance on another person; to publish a pornographic movie featuring a celebrity, or to blackmail a person by displaying a video or picture that has been changed or manipulated. Deepfakes are also used in the production of fake films depicting politicians for the purpose of fabricating news. In a nutshell, Deepfake has developed into a significant concern in today's society\cite{chen2021magdr}.
\section{ STATEMENT OF PROBLEM }
Deepfakes are becoming an urgent and visible danger to the integrity of the multimedia information we have at our disposal. Deepfakes, for example, when used on politicians and fed with targeted disinformation, might have a detrimental influence on people's opinions and have unintended consequences, including corrupted and tampered votes.
\vspace{2mm}
\\As per the Washington Post, face photos, as well as pornographic images, are now masterfully combined and transmitted through social media without the accord or understanding of the involved entities. The crime victims of these Deepfakes pictures are also spread to include the wider populace. Even some companies specialize in shipping Deepfakes. Deepfakes include pornographic images and videos where a person's face may be replaced with another person's using neural networks. Since Deepfakes are an issue for the general population, procedures for spotting them must be created \cite{dolhansky2020deepfake}. In the creation of Deepfake, tampering is not completely successful leading to certain anomalies in face structure like an anomaly in eye movement like blinking or eyebrow movement. With this review paper, we aim to study these Deepfake detection methods that employ biological features in detection. Broadly, we study five biological features: Eyebrow, eye blinking, eye movement, ear and mouth movement and heartbeat detection.
\section{DEEPFAKE}
\subsection{Deepfake Creation}

\begin{figure}[H]
    \centering
    \includegraphics{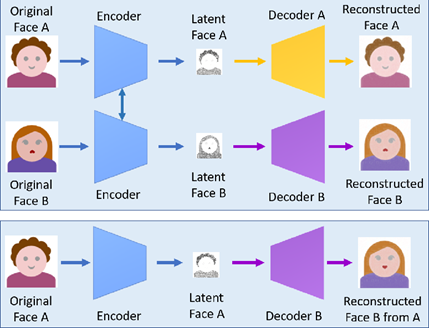}
    \caption{Deepfake Creation\cite{nguyen2022deep}}
    \label{fig:figure2}
\end{figure}
 
Due to the high calibre of the altered videos and the ease of various implementations for a broad variety of users with different computer abilities, from experts to novices, Deepfakes have grown in popularity. These applications are usually made using deep learning methods. The ability of deep learning to represent complicated and multidimensional data has a long history\cite{jung2020deepvision}. Deep autoencoders, a type of convolutional model having this feature, have been extensively used for image compression and picture compression. A Reddit user created Fakeapp utilized an autoencoder-decoder coupling architecture as the first effort to build a Deepfake\cite{abu2022analysis}. Using the latent features that the auto-encoder has gleaned from the face pictures, the decode recreates the facial image in this manner. To switch features among the target and source photos, two encoder-decoder pairings are required, each of which is learned on a collection of images, and the settings of the encoding are exchanged by the two networking pairs\cite{qi2020deeprhythm}. So, two pairs make use of the identical encoder. The common encoder can identify and understand the similarities among multiple groups of face pictures using this approach since faces have common characteristics including the locations of the eyebrows, lips, or mouths\cite{lee2022bznet}.
\vspace{3mm}
\\Large number of photos of the two subjects are used to create Deepfakes, which are created using the following steps:
\vspace{3mm}
\\Step 1: All of the photographs are first taken and encoded using a deep learning CNN by an encoder. The encoded data is then decoded using a decoder to recreate the picture. The encoder and the decoder struggle to store all of the parameters in their separate memory areas due to the enormous number of parameters \cite{sun2022faketracer}. The encoder will only remove major components that are necessary for reconstructing the original input in order to get around this issue. The decoder will continue to decode those features when the feature extraction operation is complete. Backpropagation will be used throughout the rest of the training procedure until the output and the input are identical. Graphical processing units, sometimes known as GPUs, are used since the procedure requires a significant amount of time.
\begin{figure}[H]
    \centering
    \includegraphics{{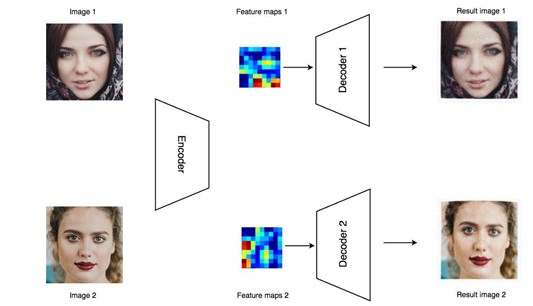}}
    \caption{Using two separate decoders\cite{nguyen2022deep}}
    \label{fig:figure3}
\end{figure}
Step 2: It involves replacing the person's visage with another's on a frame-by-frame basis once the training procedure has been completed\cite{nguyen2022deep}. Face recognition software is used to isolate person A's visage, which is then supplied into the encoder. Instead of giving it to its original decoder, the picture is fed into the decoder of person B, who then uses it to reconstruct the image. As a result of this, characteristics of individual A from the original drawings from the video are done on person B. As can be seen in the figure that follows, after this step has been completed, the freshly fabricated face is merged with the initial picture.
\begin{figure}[H]
    \centering
    \includegraphics{{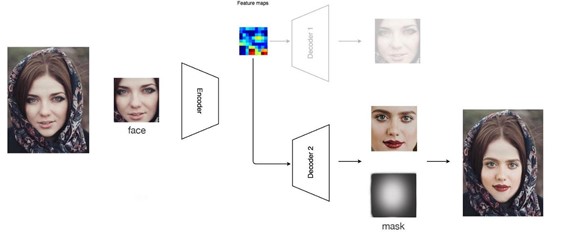}}
    \caption{Merging of Newly Created Face\cite{nguyen2022deep}}
    \label{fig:figure4}
\end{figure}
Step 3: Prior to beginning the training process, it is necessary to have hundreds or even thousands of photographs of both individuals\cite{guarnera2020deepfake}. The quality of these face photographs may be highly increased, that allows for much superior outcomes. It is essential to get rid of any pictures that have poor lighting or quality, as well as any others that include other people.
\begin{figure}[H]
    \centering
    \includegraphics{{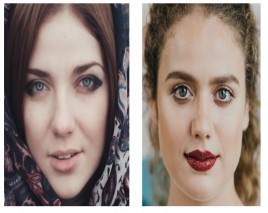}}
    \caption{Similar resemblance such as face shape \cite{nguyen2022deep}}
    \label{fig:figure5}
\end{figure}
Step 4: If the resolution of the finished image is different from the resolution of the source photo, the Deepfake will seem unnatural. This problem may be solved by cropping and rearranging the picture such that it has dimensions of 256 by 256. The training procedure is the only time the 160x160 pixel section in the middle is used; after training, it is scaled down even more to 64x64 pixels. As a consequence of this, the rebuilt faces have a resolution of 64 by 64 pixels, and they are included into a movie\cite{frank2021wavefake}. The freshly produced photos are then changed so that their dimensions are brought back in line with the originals, as seen in the figure below. However, as a result of this transition, the face will take on a hazy appearance. Both training a neural network that is able to function with pictures that have higher pixel sizes and decreasing the resolution of the video or image that is to be switched are two options that might be examined in order to circumvent this challenge.
\begin{figure}[H]
    \centering
    \includegraphics{{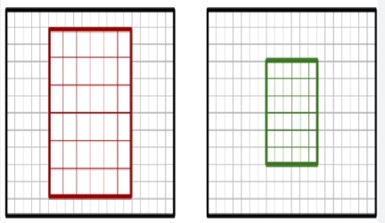}}
    \caption{Resolution morphed back to match original size\cite{nguyen2022deep}}
    \label{fig:figure6}
\end{figure}
After the Deepfake video has been made, there are two technical considerations that need to be taken into consideration as the fifth step. Both of these technical characteristics are part of artificial intelligence systems; the first of these is referred to as a discriminator, while the second of these is called a generator. Both of these systems operate in tandem with one another; the generator is the component that is responsible for producing phoney films or photographs. After the phoney video has been prepared, the discriminator will evaluate it to decide whether or not it is a fake or a real video. While it is determined that the video is a fake, the discriminator provides a hint to the generator about the preventative measures that it needs to take when it is making the subsequent clip. This results in the formation of a network known as a Generative Adversarial Network, or GAN. In order for a GAN to function, the user must first provide the desired kind of output to the system, which then generates a dataset for the generator to learn from. The generator will begin producing a variety of movies, and as soon as the necessary output has been reached, the videos will be sent to the discriminator to be analyzed. The more errors that are discovered in phoney videos, the more improvements that may be made to the generator for the following iteration. Therefore, improving the discriminator's ability to recognise false movies will also improve the generator's ability to create convincing fake videos, as the figure below demonstrates.
\begin{figure}[H]
    \centering
    \includegraphics{{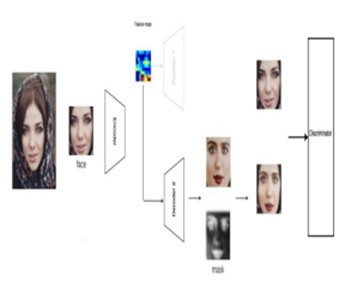}}
    \caption{Discriminator detecting whether the image is real or fake\cite{nguyen2022deep}}
    \label{fig:figure7}
\end{figure}
\subsection{Deepfake Detection}
The threat of Deepfakes to democracy, social security, and privacy is growing. Early attempts relied on fabricated traits that were the result of mistakes and shortcomings in the fake video generation process\cite{hernandez2020deepfakeson}. On the other extreme, deep learning is currently becoming used to extract features with significant and distinctive qualities in order to identify Deepfakes.
\vspace{3mm}
\\Classifiers are used  to differentiate between authentic movies and ones that have been manipulated in the process of Deepfake detection. This issue is also referred to as a binary classification problem. To train categorization algorithms using this strategy, a sizable collection of real and fraudulent films is needed\cite{lyu2020deepfake}. False movies are more accessible than ever, but their application is still restricted in terms of providing a standard for comparing different detection algorithms\cite{nguyen2022deep}. Employing clips from the searchable database, low and high quality Deepfake videos were created that accurately mimicked facial emotions, lips movement, especially eye blinking. Then, several Deepfake detection algorithms were tested using these videos. 
\begin{figure}[H]
    \centering
    \includegraphics{{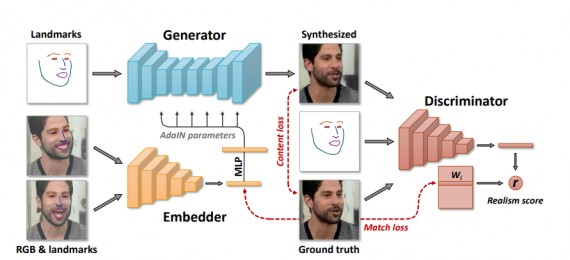}}
    \caption{Deepfake detection image}
    \label{fig:figure8}
    \cite{https://doi.org/10.48550/arxiv.1905.08233}
\end{figure}
The experiments' findings show that popular face recognition algorithms based on VGG and Face net are ineffective at effectively spotting Deepfakes\cite{lyu2020deepfake}. Whenever applied to the same freshly collected data set, additional approaches, such as the lip-syncing approach as well as SVM  picture quality evaluations, focus on providing extremely significant error rates. This elevates questions regarding the urgent requirement for the creation of future, higher reliable techniques for identifying real Deepfakes versus Deepfakes. 
\subsection{Fake Image Detection}
Face swapping offers a wide range of exciting uses, such as portrait transformation, video compositing, and, most significantly, individuality defense since it could substitute people's faces in photos with stock photos. To get unauthorized access to identification or authentication systems, cybercriminals also utilize this tactic. Since deep learning models can preserve the location, facial expression, and lighting of the shots, switched face photos are becoming more challenging for forensics algorithms. Deep learning models include CNN and GAN.
\vspace{3mm}
\\SVM, random forest (RF), and multi-layer perceptrons were a few of the classifiers used to distinguish swapped face photos from real ones. A collection of condensed features was extracted using the bag of words approach (MLP). Considering that GAN-generated pictures are the most realistic and of the highest quality among deep learning-generated images, this is likely because GAN is capable of understanding the distribution of complicated input data and producing new outcomes featuring a similar feature distribution. The generality capability of the detection methods also isn't considered in the bulk of works on the detection of GAN-produced images, notwithstanding the reality that GAN research is ongoing and several novel GAN extensions are consistently created an image pre-processing procedures to eliminate low-level, high-frequency suggestions in GAN pictures, including Gaussian blur as well as Gaussian noise. In comparison to earlier image forensics techniques or image steganalysis networks, this increases the statistical resemblance between actual and false pictures at the pixel level and drives the forensic classifier to understand additional intrinsic as well as relevant characteristics.
\subsection{Current Deepfake Detection Methods}
Face-swapping movies, which make up the overwhelming bulk of Deepfake films published online, are primarily targeted by Deepfake detection algorithms now in use. Numerous approaches now in use are founded on issues concerning binary classification at the segmented stage. Various techniques are categorized into three main groups depending on the traits they use. The initial individual categories strategies are centered on variations in the physiological as well as physical properties of the Deepfake videos. The method in use is based on the discovery that a lot of Deepfake videos lack comforting eye blinking as a result of the use of online portraits as training data, which often do not contain closed eyes for aesthetic reasons. Incoherent head movements in Deepfake films are used to uncover Deepfake movies. The second group of Deepfake detection techniques takes the use of synthesis-related artifacts at the signal level. Many methods currently in use are based on problems with binary classification at the subcellular level. According to the characteristics they use, different approaches are divided into three primary groupings. The first category-specific tactics are based on differences in the physiological and physical characteristics of Deepfake films.
\subsection{Deepfake Detection: Current challenges and Next steps}
The current issues that are now being faced as well as those that will be faced in the future with Deepfake detection Fundamentally speaking, there are three primary categories of Deepfake videos.
\begin{itemize}
\item Head puppetry involves mixing a video of a source person's head with a video of a target person's whole head as well as upper shoulders such that the synthesizing targeting seems to work in the same method as the source. This creates the illusion that the source is the target.
\item Lip syncing is the method of making a fake video by simply employing the target's lip area such that the target seems to say approximately that they do not pronounce in real life.
\end{itemize}
\subsection*{Existing challenges:}
In the meantime, the subject of identifying Deepfake films is often presented, addressed, and estimated as a binary classification problem. As a result, every video is classified as either authentic or a Deepfake. When developing and testing Deepfake detection algorithms employing movies that are neither original nor made employing Deepfake production methods, it is feasible to establish a contradiction of this kind. However, when the detection approach is employed in real-world scenarios, the issue becomes more challenging.
The circumstance that a video is not recognized as a Deepfake does not always imply that the video is a legitimate one. For instance, movies may be produced or altered using techniques other than Deepfakes. A single label would not be able to adequately represent the fact that a Deepfake clip may have undergone different alterations. Additionally, Deepfake will only create one or a small number of those faces for a fraction of the frames in the movie if a film comprises the faces of many different persons. Therefore, to fully handle the complexity of real-world media forgeries, the binary classification approach has to be enhanced to incorporate multiclass, multilabel, and local classification and detection.
\section{RELATED WORK}
In the research that has been done so far, a variety of methods for detecting Deepfake photos and videos have been suggested. a comparison of the most important techniques in the field, with a focus on the ones that are based on physiological monitoring and paying particular attention to the bogus detectors. We provide information on the previous study's methodology, as well as its classifiers, highest achievable level of performance, and research databases.
\subsection{Current Directions in Deepfake Studies}
The first proposed GANs model \cite{goodfellow2014advances} is notable because it developed a brand-new technique for learning by gathering information with the Generator and authorizing it using the Discriminator. The saddles' difficulty and the value function problem \cite{radford2015unsupervised}, which cause strange ranges in the shape and shade of produced images, are two defects that it does have. A research team from the University of Washington built upon this improvement in 2017. They created complex false movies that replicated the voice and mouth of a speaker in a video and generated the exact form of his mouth at every instant \cite{suwajanakorn2017synthesizing}. They achieved this by using university-developed software. This allowed the adoption of methods like jaw correction\cite{bayar2016deep} to greatly improve upon the prior limits of pixel crush, jaw form, wrinkles, and so forth.
\vspace{3mm}
\\Alec Radford and colleagues presented (DCGAN) Deep Convolutional Generative Adversarial Networks in 2016, which allowed for the computation of arithmetic operations with filtering among pictures employing latent vectors. This was done by using CNN models on GANs, which produced more advanced fakes \cite{radford2015unsupervised}.
\subsection{Research on Detection of Deepfakes}
For generalized fake video detection \cite{cozzolino2017recasting}, face matching \cite{rossler2018faceforensics}\cite{hsu2018learning}, and eye blinking [42], they looked at a variety of forensic techniques. The most popular Deepfake detection techniques include training deep neural networks on a dataset of forged faces or spotting an aberrant pixel \cite{guera2018deepfake}\cite{koopman2018detection}.
Many Deepfakes faces, according to a study \cite{guera2018deepfake}, do not blink their eyes. However, a few more recent instances have modified the discriminator to examine blinking to circumvent these detection techniques\cite{rossler2018faceforensics}.
\vspace{3mm}
\\To develop deep learning-based algorithms, Face Forensics++ \cite{rossler2018faceforensics} is a helpful dataset of face forgeries. Additionally, research \cite{hsu2018learning} describes a method for spotting fake videos using a trained CNN. The techniques produce reliable findings, but they necessitate a large amount of input and frequent updates. As a consequence, we focused on research, which requires less information but is more effective. It will probably be used more widely\cite{lawrence1997face}.
\vspace{3mm}
\\The literature has also looked at methods based on the identification of face-warping artifacts. Among the best ways to spot concealed face modifications is through CNN-based identification algorithms, which can recognize the presence of such abnormalities from the face as well as its surroundings\cite{li2020sharp}.
\vspace{3mm}
\\The most widely used fake detectors are probably those that use just deep learning features, putting as many actual and fake videos into the networks as feasible and letting the networks automatically extract the distinguishing traits. These false detectors have generally proven effective. outstanding results when using standard network architectures.\cite{cozzolino2021id}
\vspace{3mm}
\\Using cutting-edge network topologies, the scientists assessed each face region's capacity for discrimination and obtained intriguing findings on Deepfake datasets from the first and second generations\cite{tolosana2001beyond}. This is in contrast to the majority of methods, which rely on the complete face to identify fraudulent movies.
\vspace{3mm}
\\Table 1 highlights the research publications on Deepfake detection and Super-resolution, as well as the various strategies utilized for Deepfake detection. The following table analyzes the different techniques and characteristics utilized for Deepfake detection. It comprises approaches based on Machine Learning and Deep Learning.

\begin{table}
 \caption{Analysis Table}
 \centering
  \begin{tabular}{|c|l|l|l|c|}
    \toprule
    \textbf{Sr. No.}& \textbf{Title of Paper}& \textbf{Techniques used}&\textbf{Dataset used} &\textbf{Accuracy}\\
    \midrule
    \hline &&&&\\
    \cite{yadav2019deepfake} &  Deepfake: A Survey on  & 1. Convolutional Neural  &Face2Face, Reddi  &  {95\%} \\
    &   Facial Forgery Technique & \hspace{0.3cm} Networks (CNN) &  user Deepfakes& \\
    &   Using Generative Adversarial &  2. Long Short-Term & &\\
    &   Network &\hspace{0.3cm}Memory (LSTM)& &\\
    &&&&\\\hline &&&&\\
   \cite{amerini2019deepfake} &  Deepfake Video Detection  & Convolutional Neural  &Face2Face &  VGG16 {81.61\%},   \\
    &   through Optical Flow &   Networks (CNN) &  & ResNet50 \\
    &  based CNN &  & &{75.46\%} \\
    &&&&\\\hline &&&&\\
   \cite{guera2018deepfake} &  Deepfake Video Detection  & 1. Convolution Neural  &HOHA dataset &  Conv-LSTM  \\
    &   Using Recurrent Neural  & \hspace{0.3cm} Networks (CNN) &  &(20 frames)  {96.7\%}, \\
    &   Network &  2. Long Short-Term & &Conv-LSTM\\
    &    &\hspace{0.3cm} Memory (LSTM)& &(40 frames)\\
    &&&&\\\hline &&&&\\
   \cite{ranjan2020improved} & Improved Generalizability of   & 1. Convolutional Neural  &1. FaceForensics++   &  With Transfer \\
    &  Deep-Fakes Detection  &\hspace{0.3cm}Networks (CNN) &  2. Celeb-DF &Learning {84\%}, \\
    &  Using Transfer Learning &  2. Long Short-Term &3. Deepfake Detection& Without Transfer \\
    &   Based CNN Framework &\hspace{0.3cm}Memory (LSTM)& Challenge&Learning {75\%}\\
    &&&&\\\hline &&&&\\
   \cite{luo2018multi} &  Multi-scale face   & 1.Convolution Neural  &1. Celeb A   &  Discrete- {95\%} and \\
    &   detection based on &\hspace{0.3cm}Networks (CNN) &  2. AFW & for continuous, it \\
    &  convolutional neural  &  & 3. FDDB &  is {74\%}\\
    &   networks.& & & \\
    &&&&\\\hline &&&&\\
   \cite{matern2019exploiting} &  Exploiting Visual Artifacts  & The neural network  &1.CelebA   &  MLP {84\%(Eyes)},\\
    &   to Expose Deepfakes  &  classifier as MLP and & 2. ProGAN & LogReg \\
    & and Face Manipulations & the logistic regression & 3.Glow & {83\%(Eyes)}\\
    & & model as LogReg & &\\
    &&&&\\\hline &&&&\\
    \cite{mccloskey2019detecting} & Detecting Gan-generated & SVM classifier  & Image net dataset      &   {92\%,}  \\
    & imagery using saturation cues &&&\\
    &&&&\\\hline &&&&\\
    \cite{kharbat2019image} & Image Feature Detectors & 1. SVM classifier & Unnamed with & HOG {94.5\%},\\
     & for Deepfake Video Detection & 2. Feature extraction & 98 videos & SURF {90\%},\\
     &&\hspace{0.3cm}algorithms & KAZE {76.5\%}&\\
    &&&&\\\hline &&&&\\
    \cite{malolan2020explainable} & Explainable Deep-Fake & 1.Xception net (CNN) & Face Forensics++ & {90.17\%}\\
    & Detection Using Visual & 2.LRP and LIME &&\\
    & Interpretability Method &&&\\
&&&&\\
    \bottomrule
  \end{tabular}
  \label{tab:table}
\end{table}

\section{PERFORMANCE ANALYSIS}
Currently, Deepfake video identification is frequently discussed, analyzed, and presented as a binary classification issue. Each video is thus categorized as either legitimate or a Deepfake. In randomized studies wherein we design as well as test Deepfake detection techniques employing movies that are neither authentic nor made utilizing Deepfake production techniques, it is simple to set up a dichotomy similar to the ones indicated above. When the detecting approach is used in the actual world, the situation becomes murkier. Different than using Deepfakes, videos might well be created or edited in rather ways\cite{korshunov2019vulnerability}, thus just because a video isn't classified as a Deepfake doesn't always mean it's a genuine one. A single label may not sufficiently convey this if it occurs that a Deepfake video has been subjected to numerous forms of manipulation. Additionally, Deepfake will only create one or a limited number of faces in a video with many different people's faces for a tiny portion of the frames. The development of the binary classification method to include multi-class, multi-label, including local categorization as well as identification is necessary due to the complexity of real-world media forgeries.
\vspace{3mm}
\\The capacity of deep learning techniques, such as CNN as well as GAN, to maintain picture position, facial expression, as well as illumination, has made it more demanding for forensics algorithms to distinguish amongst swapped face shots \cite{zhang2017automated}. Zhang et al\cite{pu2020noisescope}'s bag of words approach was utilized to extract a set of condensed characteristics, that were then passed into a variety of classifications, including SVM, RF, and MLP, which are two techniques that can distinguish between images of swapped faces and real faces. Because they are the most realistic and high-quality pictures produced by deep learning, those produced by GAN models are arguably the most difficult to distinguish. This is so that fresh outputs produced by GAN models will have a distribution that is comparable to the learned distribution of the complicated input data.
\vspace{3mm}
\\Regardless of the fact because GAN is continuously developed and several new extensions are often published, the majority of research on the detection of GAN-generated images does not examine the generalization capacity of the detection models. Even if GAN is often extended in novel ways, this is still true. For instance, Xuan et al. \cite{yang2021defending} used Gaussian blur and Gaussian noise to GAN pictures to filter out low-level, high-frequency information. The forensic classifier must acquire more intrinsic and relevant traits due to the rising pixel-level statistical similarity between real and fake pictures.
\vspace{3mm}
\\On the other side, a problem with hypothesis testing is the GAN-based Deepfake detection. They did this by introducing a statistical framework by making use of an information-theoretic analysis of authentication. Oracle error is the term used to refer to the smallest acceptable gap that can be established between the distributions of real pictures and those of images produced by a certain GAN.
\section{EXPLAINABILITY}
Explainability is another difficult aspect. In many different situations, a numerical score indicates the possibility that the picture or video was fake. In situations like this, it is an extremely common practice to inquire about the reason behind the numerical score that must be acceptable for the analysis. Moreover, several data-driven DF detection technologies are black boxes. Therefore it is difficult for these algorithms to explain their results. This is especially true about techniques that rely on deep neural networks model DNN. Additionally, because of the "black box" character of DNN algorithms, many data-driven DF detection systems, particularly many which rely on their use, cannot usually understand the results.
\vspace{3mm}
\\\textbf{Temporal Aggregation:}The majority of the currently available techniques for detecting Deepfakes at the segment level are predicated on binary identification. In other words, they estimate the likelihood that a certain frame is Deepfake or real. Temporal aggregation, meanwhile, goes one step further. Although it seems simple and uncomplicated, this approach has two significant drawbacks. Firstly, even though several Deepfake films include behavioral abnormalities, real or Deepfake images generally appear at constant intervals. The chronological integrity across pictures also isn't specifically checked. And secondly, it requires an additional step to be taken whenever a score for the video's integrity is needed. To calculate such a score, we must first aggregate the scores obtained from each frame.
\vspace{3mm}
\\textbf{Protection measures:} One of the many approaches they have recently researched is to include adversarial perturbations, which are carefully created patterns that are undetectable to human eyes but may result in detection failures. This is one of the numerous methods they have recently explored. The second element, which involves making a genuine video that seems like a Deepfake video by incorporating approximated transmitted signal characteristics required by existing detection algorithms, could also be employed to construct anti-forensic techniques. This situation is what we call phoney Deepfake. In this regard, anti-forensic methods may also be devised. It is necessary to enhance further Deepfake technology detection to counter both harmful and intentional attacks.
\vspace{3mm}
\\\textbf{Social Media Laundering:} The dissemination of a sizable amount of internet videos is rapidly being done via social media sites like Facebook, Instagram, and Twitter. These videos often undergo extensive compression, meta-data removal, and file size reduction before being uploaded to social media. This is carried out to protect viewers' privacy as well as to reduce network traffic. These actions, more often known as social media laundering, make it harder to identify underlying manipulation and increase the chance of false positive detections, which is the practice of categorizing a real video as a Deepfake.
\vspace{3mm}
\\\textbf{Anti-forensics:} The second element, entails introducing simulated signal characteristics needed by current detection techniques to make real footage appear to be Deepfake footage, which may also be used to construct anti-forensic techniques. We refer to this circumstance as a false Deepfake. Anti-forensics: The second element allows for the development of anti-forensic measures. Additional Deepfake detection algorithms need to be improved to deal with these purposeful and antagonistic attacks.
\vspace{3mm}
\\However, biological features such as eyebrow recognition, eye blinking detection, eye movement detection, ear and mouth detection, and heartbeat detection are clearly described. A comparison of each biological feature with different classifiers/techniques along with its key findings is discussed.
\section{BIOLOGICAL FEATURES}
\subsection{Eyebrow Recognition}
In \cite{nguyen2020eyebrow}, they used eyebrows to identify Deepfakes via biometric comparison. Deepfake detection method has used a biometric comparison of the brow region. One of the most affected areas in the synthesized images is the brow region. When examined by a biometric comparison pipeline, eyebrow alterations become more recognizable, particularly in high-resolution and high-quality Deepfakes. The model must first discover the participants' identity for this method to be effective (biometric enrolment is needed). Additionally, this will hold if the targets are well-known individuals like politicians or celebrities. Examples include LightCNN, Resnet, DenseNet, and SqueezeNet. In the publication of biometric research, they are often used. They believe they will do well in the brow-matching task as a consequence.
\vspace{3mm}
\\A large and excellent forensic dataset was created for Celeb-DF. This dataset contains 5639 Deepfake films in addition to 590 original footage from 59 celebrities. The paper has analyzed the highest quality Deepfake dataset, which provides the most convincing fake video following image synthesis visual artifacts. Although this is a challenging task, it could be useful in outlining the advantages of the recommended technique. Celeb-DF, in contrast to the other datasets, mostly lacks apparent Deepfake artifacts such as splicing boundaries, color mismatches, and inconsistent face orientation. As a result, several studies on this dataset's Deepfake detection have claimed low accuracy ratings.
\subsection{Eye Blinking Detection}
In \cite{li2018ictu} , the aim is to find facial videos created by AI by looking for the lack of eye blinking. To represent the phenomenology and chronological recurring patterns in the operation of eye blinking, CNN and RNN are combined in a unique deep-learning algorithm. To classify between open-eye and closed-eye states in a frame, current methods use Convolutional Neural Networks (CNN) as a binary classifier. CNN, on the other hand, bases its predictions exclusively on a single frame and ignores temporal domain data.
\vspace{3mm}
\\Long-term Recurrent Convolutional Neural Networks (LRCN) are employed to classify between open and closed eye states while taking past temporal information into account since human eye blinking has a considerable temporal relationship with earlier states. The proposed method in the literature performs well in identifying Deepfake videos when tested against benchmark datasets for eye-blinking detection.
\vspace{3mm}
\\A method is  suggested for spotting eye blinking in clips and the general pipeline of the method discussed is given as follows. To discount head movements and variations in alignment utilising facial landmark point recognition, the recommended approach first detects faces in each video frame. The detected faces are then aligned into the same coordinate system. The sections matching each eye are eliminated to create a stable sequence. After completing these pre-processing steps, the LRCN (Long-term Recurrent Convolutional Networks) model is utilized to detect eye blinking by determining how wide an eye is open during each video frame.
\subsection{Eye Movement Detection}
FakeET \cite{gupta2020eyes} is an eye-tracking dataset used to study how Deepfake videos are perceived by viewers. Considering that Deepfakes' primary objective is to deceive human observers, FakeET was developed to understand and gauge viewers' propensity to spot artificial video artifacts. It is possible to measure the differences between exploratory and explanation by analyzing the waveform using a variety of eye track-based measurement techniques, including preoccupation intensity, scan-path length, reaffirmed, and fixation entropy \cite{li2021exposing}.
\vspace{3mm}
\\FakeET is the name of the initial set of user behavioral responses for Deepfake detection. Via extensive experimental investigation, they show the value of both the eye-gaze and EEG modalities in enabling automated Deepfake recognition. Both the eye gaze and the EEG in \cite{gupta2020eyes} modalities reveal distinct patterns suggestive of information problems. Google/Jigsaw Deepfake Detection is the database. 811 films—331 genuine and 480 fake—were included in our study out of the 3068 false and 363 real videos in the collection. Furthermore, based on the visual inspection of the fake videos, they were categorized into easy and difficult forgeries; this categorization yielded 244 simple and 236 difficult-to-detect phoney videos\cite{masood2022deepfakes}.
\vspace{3mm}
\\Based on the reported findings in \cite{gupta2020eyes}, the Dec-Tree and T-CNN algorithms outperform random in terms of F1-score, the k-NN classification obtains the second-highest F1 metric of 0.54, and the Naive Bayes classifier produces the best EEG-based FD having AUC and F1 ratings of 0.55. A random classifier performs similarly to regression models and linear SVM. While a small increase in FD results was achieved in this study, stronger EEG descriptors (such as Spectral Density characteristics and EEG spectrograms) and better classification techniques may be investigated for potential advancements\cite{startsev2022evaluating}. Generally, the results of EEG-based categorization show that Better-than-chance FD can only be achieved using PCA-ed EEG information and incorrect classifying techniques.
\subsection{Ear and Mouth Movement}
In a lip-sync Deepfake, a person’s mouth area is swapped with a fabricated or artificially manufactured audio recording. The human ear was not taken into consideration while making these deep-fake films. The ears belong to the original self even if the face identification in a face swap Deepfake might exactly resemble the co-opted identity. Even if the lips may be perfectly synchronized to the music in a lip-sync Deepfake, the ear's motion characteristics will be separated from the mouth as well as jaw movements \cite{agarwal2021detecting}.
\vspace{3mm}
\\For the above research, a total of 64 films of Mark Zuckerberg, Donald Trump, Angela Merkel, and Joe Biden were downloaded from YouTube. The duration of these films ranged from 12 to 59 seconds. In each video, the left or right ear was audible. They calculated the audio RMSE, the vertical separation between the lips, and the auditory motion for each frame of each video. In a lip-sync Deepfake, an existing video’s mouth movement is altered to relate to a new soundtrack \cite{gu2022delving}.
\vspace{3mm}
\\The correlations of face, aural and auditory signals across all Biden, Merkel , Trump and Zuckerberg- starring clips are determined between the facial (lip vertical) or auditory (RMSE) signal and the helix, tragus, or lobules horizontal and vertical motions. The connections are mostly characterized by a person-specific nature. Trump, but not the other candidates, displays a strong negative correlation, similar to the vertical lobule motion. The vertical aural motion follows this core structure consistently. For Trump, but not for remaining candidates, the tragus movements are substantially associated with audio. A linear classifier is trained in the manner described below to determine how effectively these dynamical acoustic parameters can detect lip-sync Deepfakes. For the accessible films, every person's instruction and assessment sets are classified into non {80\% / 20\%} groups. Using 12 aural/oral associations between the real videos and the simulated false ones, a multinomial logit model is developed. Then, utilizing real movies together with simulation, Generative adversarial network, and in-the-wild false films, this model was assessed\cite{elhassan2022dft}.
\vspace{3mm}
\\The mean evaluation accuracy rises from 0.84 to 0.87 with this individual training as well as the mean mentoring AUC rises from 0.91 to 0.98. Despite just looking at brief (10-second) portions, accuracy is often rather good. A complete film may be integrated with a simple majority rule to increase this accuracy\cite{gerstner2022detecting}.
\subsection{Heartbeat Detection}
For deepfake detection that used heartbeat, a research paper, \cite{boccignone2022deepfakes} that leverages rPPG and heartbeat was reviewed. The suggested deepfake detection method based on the physiological heart rate estimate postulates that when calculating the localized pulse rate from a picture of a genuine head vs. a counterfeit one, rPPG (remote photoplethysmography)  approaches should yield noticeably different findings. In their proposed method, after detecting the face of the (susceptible) manipulated subject, a group of 100 patches is automatically tracked on it. The pixels color intensities within time are averaged, thus forming 100 RGB traces. Further, for each patch, blood Volume Pulse is quoted. Disruption is anticipated in each patch with BVP signals. Thus, certain features analyzed pertain to have a clear meaning as to such physiological behavior which lead to DeepFake detection otherwise . In \cite{fernandes2019predicting}, it is believed that the majority of the current Deepfake altering algorithms do not yet account for the physiological characteristics of a human being since the color changes and lighting induced by oxygen content are slight and imperceptible to the human eye. The early design of Deepfakes ON Phys was based on the Big Phys model, whose goal was to estimate the actual heart rate using face video sequences\cite{fernandes2019predicting}. The deep learning-based model's objective was to collect temporal data from movies in a manner resembling the behavior of classic manual rPPG procedures. It is possible to extract facial traits by observing the change in color in individuals' faces, that are brought on by variations in blood oxygen levels. Additionally, signal processing techniques are employed to differentiate between color changes brought on by plasma and other alterations that may be brought on by noise and ambient light.
\vspace{3mm}
\\A thorough experimental evaluation of DeepFakesON-Phys uses Celeb-DF v2 and DFDC Preview, two of the most current available datasets from the second Deepfake generation. Google created one of the biggest public databases accessible via partnerships with other corporations and major universities including Microsoft, Amazon, as well as MIT. The name DFDC appears in the database. The 1,131 genuine recordings made by 66 performers who were contracted for the DFDC Preview collection provide realistic diversity in genders, skin color, and age. Celeb-DF v2 is one of the most difficult Deepfake datasets to use at the present, and it is important to emphasize that, unlike other well-known databases, this collection was not produced using either publicly accessible information or data from social media platforms. The Celeb-DF v2 database's goal was to produce fake videos with superior graphics than those in the UADFV database. This collection includes 590 actual YouTube videos on well-known individuals who identify with various racial, ethnic, or gender identities.
\vspace{3mm}
\\The recommended method analyses both short-term video level and frame-level fake detection while assessing DeepFakesON-Phys. In both tests, DeepFakesON-Phys outperforms the state-of-the-art and produces amazingly accurate answers.
\vspace{3mm}
\\Table 2 represents the biological feature and classifiers / techniques, and used data sets, key findings of biological features.
\begin{table}
 \caption{Comparison of biological features}
 \centering
  \begin{tabular}{|c|l|l|l|l|l|}
    \toprule
   \textbf{Ref} &\textbf{ Biological}  &  \textbf{Classifiers/ Techniques} & \textbf{Datasets Used} & \textbf{Dealing} & \textbf{Key Findings}\\
     &\textbf{ feature}& & \textbf{ with}& &\\
    \midrule
    \hline&&&&&\\
    .\cite{mao2021exposing} & Heart Rate &Photoplethysmography & Face & Images & The plan increases overall\\
    & & and autoregressive & Forensic ++ & & accuracy for the Face\\
    & &     (AR) model & dataset, Celeb-DF & &  Forensic ++ dataset by \\
    & & & dataset & & {25.66\%} when associated with\\
    & & & & &  the baseline network and by\\
    & & & & & {1.48\%} when compared to the\\
    & & & & &  baseline approach.\\
    &&&&&\\\hline &&&&&\\
    .\cite{li2018ictu} & Eye &LRCN method&CEW Dataset&Images/&This approach is evaluated\\
    &movement&&&Videos& using benchmark datasets\\
    &&&&& for eye-blinking detection,\\
    &&&&& and it exhibits good results\\
    &&&&&  when used to identify films\\
    &&&&&   produced using Deepfake.\\
    &&&&&\\\hline &&&&&\\
    .\cite{nguyen2020eyebrow} & Eyebrow  &LightCNN, ResNet, & Celeb-DF & Images & Because once implemented to \\
    & movement & DenseNet, SqueezeNet & [Li20] & & the best quality Deepfake\\
    &&&&&  dataset, our technique offered\\
    &&&&&  a 0.88 AUC and 20.7\% EER\\
    &&&&&   for Deepfake identification.\\
    &&&&&\\\hline &&&&&\\
    .\cite{agarwal2021detecting}&Ear and&Aural dynamics&64 videos were&Videos&The measured signal unfolds\\
    &mouth& &downloaded& & across hundreds of frames\\
    &movement& &from YouTube of& & in dynamic auditory\\
    & & &Joe Biden,&& and oral analysis, in contrast\\
    & & &Angela Merkel,& & to existing synthesis approaches,\\
    & & &Donald Trump,& & which generally work on\\
    & & &and Mark & &  one or a small number\\
    & & &Zuckerberg.& & of video frames. \\
    &&&&&\\\hline &&&&&\\
    
    .\cite{gazi2021deepfake}&  Eyelid & LRCN method, & UBIRIS, & Images & In terms of eyelid aperture\\
    .\cite{fuhl2016eyes} & & computer vision-based & CASIA & & estimate and eyelid outline\\
    & &  approach &   & & similarity, the suggested \\
    & & & & & technique performs better than\\
    & & & & &  the state-of-the-art.\\

    &&&&&\\\hline &&&&&\\
    .\cite{jung2020deepvision}& Eyeblink & adversarial network & Face & Images & The recommended method\\
     & & (GANs) model & Forensics & & called as Deep Vision\\
     & & & ++& & is utilized as a measurement\\
     & & & & &  to verify an abnormality\\
     & & & & &  whenever eye blinks are\\
     & & & & &  frequently replayed in a very\\ 
    & & & & &short period of time depending\\ 
    & & & & & on the period, the repeating\\ 
    & & & & &  frequency, as well as expiring \\ 
   & & & & &   eye blink length.\\
    &&&&&\\\hline &&&&&\\
    .\cite{lin2017pulse} & Pulse rate & PPG sensors & ECG signal & Images & According to the findings of \\
    &&&&& the experiment, the suggested\\
    &&&&& strategy may improve the\\
    &&&&& usefulness and likelihood of\\
    &&&&& PPG signal identification on\\
    &&&&& palms.\\

    \bottomrule
  \end{tabular}
  \label{tab:table}
\end{table}
\section{DISCUSSION}
In this paper, we mainly concentrated on both the generation and detection of Deepfakes and also existing Deepfake detection techniques and their difficulties. However, biological aspects like recognizing an eyebrow, detecting an eye blink, detecting an eye movement, detecting an ear or mouth, and detecting a pulse are precisely defined. The important findings from a comparison of each biological trait using various classifiers and methodologies are highlighted. Finding the integrity verification index values in the GANs method that are extremely challenging to evaluate requires utilizing the discriminator, producing fake pictures, a single recurring and subconscious activity, which provides a new technique. The similarity of the data generated by GANs to actual data is one of its most alluring features. They, therefore, have many different real-world uses. They can create text, pictures, audio, and video that are identical to real data. One pertinent scope was identified by studying these deepfake detection methods that use biological features for identification - can these biological features be altered by using high quality filters in video processing or by applying make-up to alter these features namely fine lines in face or eye-brow movement ? Use of these smoothing techniques can cause hindrance in identifying distinguishing features of facial regions that are leveraged by these identification methods. Thus, these methods may result in giving false positives to the video that are actually not deepfakes.

\bibliographystyle{plain}
\bibliography{references}
\end{document}